# Weakly Supervised Pathology-Informed Representation Learning for PET-Based Content Retrieval of Intra-Tumour Heterogeneity

Rajat Vashistha,[1,2] Sandra Brosda,[2] Lauren G. Aoude,[2]
Christine Jestin Hannan,[2,3] James M. Lonie,[2] Jessica Ng,[4] Andrew Nathanson,[4] Ellie Vloedmans,[2] Caroline Cooper,[4]
Andrew P. Barbour,[2,4,5,*] Viktor Vegh,[1]

[1]School of Human Movement and Nutrition Sciences, Faculty of Health, Medicine and Behavioural Sciences, The University of Queensland, Brisbane, Australia
[2]Frazer Institute, Faculty of Health, Medicine and Behavioural Sciences, The University of Queensland, Brisbane, Australia
[3]Department of Surgical Sciences, Uppsala University, Uppsala, Sweden.
[4]Princess Alexandra Hospital, Brisbane, Australia
[5]PA Southside Clinical Unit, Faculty of Health, Medicine and Behavioural Sciences. The University of Queensland, Brisbane, Australia
[*]Corresponding author: a.barbour@uq.edu.au

Abstract

We propose a weakly supervised $^{18}$F-FDG-PET representation-learning framework for content-based medical image retrieval, using H&E-derived information during training while preserving PET-only inference. The proposed method was designed to use H&E-derived information during training while maintaining PET-only inference. A teacher–student training strategy was used to learn the PET tumour derived-voxel representations, from which global and hotspot-conditioned embeddings were generated along with maps of intra-tumour heterogeneity in our oesophegeal cancer test case. A progressive ablation strategy was used to evaluate the contribution of different supervision mechanisms. Retrieval performance was assessed across cross-validation folds using metrics including mean average precision, normalised discounted cumulative gain and mean reciprocal rank. Additional analyses evaluated ablation performance, hotspot faithfulness through perturbation/deletion experiments, prototype-specific PET uptake behaviour and indirect patient-level concordance between learned PET prototype classes and selected histomic features. Progressive introduction of pathology-informed supervision and hotspot modelling improved PET retrieval performance compared with global PET representations and conventional PET baselines. Across the ablation ladder, PET hotspot-conditioned representations consistently provided stronger retrieval than global embeddings, indicating that focusing on informative tumour subregions improved sensitivity to intra-tumour heterogeneity. Histopathology concordance further showed that the learned classes were not simply high-uptake PET regions; instead, they demonstrated distinct heterogeneity in $^{18}$F-FDG uptake. Weak multimodal supervision can improve PET-based retrieval by encouraging representations that capture intra-tumour heterogeneity. The proposed framework enables PET-only inference while incorporating pathology-informed training signals, supporting its potential as a virtual-biopsy approach. The code will be made available upon publication at GitHub: *https://github.com/rajat-mae/virtual-biopsy-cbmir.git*

Highlights

- Developed a weakly supervised PET representation-learning framework for content-based retrieval in oesophageal cancer.
- Integrated H&E-derived training supervision while preserving PET-only inference through teacher–student distillation.
- Introduced hotspot-conditioned and implicit derived-voxel PET representations to model intra-tumour heterogeneity.
- Showed that learned hotspot prototypes were not simply driven by high $^{18}$F-FDG uptake.

## 1. Introduction

$^{18}$F-FDG PET scans are essential for staging, treatment planning and response assessment in cancer care by delineating the whole tumour metabolic activity [1]. However, interpretation of a PET scan is still dominated by global uptake measures reflected in image hotspots, which may not fully discern the biological diversity of the tumour microenvironment. Radiomics and deep learning approaches have gained attention as they can extract richer quantitative information for prognosis, response prediction and nodal staging in cancer from $^{18}$F-FDG uptake scans [2-4]. These approaches are trained using data on patient survival, and cancer stage or treatment response. As such, they face the risk of learning endpoint-specific associations and unlikely capture the heterogeneity of tissue biology across a PET hotspot.

Histopathology provides a complementary view of tumour biology. Haematoxylin-and-eosin (H&E) whole-slide images inform on cellular morphology, stromal organisation, immune contexture, necrosis and microarchitectural patterns. Recent advances in computational pathology have shown that whole-slide images can be combined with genomic or transcriptomic information for survival prediction and biological interpretation. For example, multimodal co-attention and pathway-based transformer models have been proposed to learn interactions between histology patches and molecular features [5, 6]. More recently, optimal-transport-based multimodal learning was used to impose global structural consistency when matching histological and genomic representations [7]. In parallel, the use of the CONCH pathology foundation model demonstrated that large-scale image–text pretraining can generate transferable H&E representations for downstream pathology tasks [8]. Particularly for the cases of weak labels, sparse and noisy datasets, these studies establish the benefit of training using multimodal datasets to improve the learning of biological representations.

To date, most multimodal oncology models remain pathology-centred or genomics-centred. They typically use H&E, transcriptomics or genomics directly at inference. Emerging radiology–pathology fusion studies suggest macroscopic imaging and microscopic tissue morphology can provide complementary information for outcome prediction [9-11]. Nevertheless, relatively little work has investigated whether histopathology and genomics-derived supervision during training can improve PET-only tumour representations at inference. This gap is important because PET is acquired non-invasively before treatment and samples the whole tumour volume, whereas H&E provides high-resolution detail of biology but is invasive, spatially selective and not generally available across the entire tumour. A clinically practical representation-learning strategy should therefore incorporate sparse H&E data for model development and predict tissue biology from $^{18}$F-FDG uptake alone.

The central methodological challenge is that PET and H&E are not naturally aligned. PET is a three-dimensional, low-resolution metabolic image acquired *in vivo*, whereas H&E is a two-dimensional, microscopic tissue image acquired *ex vivo* after sampling, fixation, sectioning and staining. Collecting the spatially registered PET and the corresponding histopathology data is not clinically feasible. Without explicit PET-to-H&E spatial correspondence, it is hard to verify that a PET voxel corresponds directly to a specific H&E patch [12]. This makes cross-modal labelling scientifically challenging. For example, a hard nearest-neighbour assignment between PET and H&E embeddings may capture patient identity, acquisition artefacts, staining variation or feature-space bias rather than genuine spatial biological heterogeneity. Therefore, any PET–H&E training strategy must control leakage, preserve PET-only inference and evaluate whether pathology-informed supervision improves downstream applications beyond PET-only baselines. This limitation motivates the present study.

Instead of assuming direct spatial correspondence between PET and H&E, we formulate the task as weak multimodal representation learning. The aim is not to replace histopathology or to claim voxel-level pathological ground truth, but to test whether H&E-informed phenotype structure can guide PET embeddings toward biologically meaningful tumour similarity. To address this, we tested distinct choices for models and loss functions. The study has the following contributions:

i. a hotspot-conditioned retrieval framework to test local evidence-aware PET embeddings against global PET representations,
ii. an optimal-transport-guided model for learning PET phenotype structure without PET–H&E spatial registration, and
iii. a validation framework to evaluate retrieval accuracy, hotspot faithfulness through pseudo-voxel deletion, and comparison against radiomics and PET-adapted deep-feature baselines.

These contributions position the PET-based hypothesis-generating pattern-recognition framework at the forefront of identifying biologically informative tumour habitats from *in vivo* PET scans.

## 2. Methods

### 2.1. Motivation to link PET with H&E

A substantial body of PET-guided stereotactic biopsy and spatial co-registration studies demonstrate that $^{18}$F-FDG uptake exhibits meaningful intra-tumour heterogeneity that reflects underlying histopathology at the regional level [13]. In high-grade gliomas, PET-guided biopsies have shown that stereotactically sampled regions with higher FDG uptake are significantly enriched for anaplastic histology, while lower-uptake regions often correspond to lower tumour cellularity, necrosis, or non-neoplastic tissue. These specific findings confirm that PET voxel signal varies systematically within tumours in a spatially resolved manner rather than representing a single homogeneous lesion. Additional region-by-region analyses across multiple biopsy samples demonstrate a statistically significant relationship between local $^{18}$F-FDG uptake and histologic grade, with metabolic indices derived from PET able to discriminate anaplastic from non-anaplastic tissue at sampled coordinates, supporting voxel-level correspondence between uptake and tumour biology.

Complementary PET-guided stereotactic biopsy studies using multiple tracers similarly report that intra-tumoural differences in tracer uptake align with local variations in anaplasia and necrosis, reinforcing the concept that $^{18}$F-FDG-PET captures spatially distributed biological heterogeneity within individual lesions [14]. More recently, quantitative autoradiography of PET-guided core biopsies has enabled millimetre-scale registration between *in situ* tracer concentration and histopathology, demonstrating a direct link between measured tracer activity in tissue specimens and PET voxel uptake (correlation ≈ 0.6; natural microheterogeneity of tissues within PET hotspots leads to elevated and non-specific uptake), while also confirming high uptake PET voxels represents a spatial mixture of viable tumour, stromal components, and necrosis rather than a single histologic entity [12]. Together, these studies provide strong evidence for voxel-level PET uptake variations within tumours correspond to underlying spatial heterogeneity in tumour cellularity, grade, and microenvironment, thereby provide support for data driven approaches to map local histopathological structure across PET hotspots.

## 2.2. Study overview

Figure 1 provides a broad overview of the teacher-student learning framework to predict tumour heterogeneity within PET hotspots. The steps in the figure have the objective of learning the PET hotspot-derived embeddings that could be associated with tissue-level tumour phenotypes in patients presenting with oesophegeal cancer. The H&E information was used only as weak supervisory evidence during training, and it was omitted during inference (i.e., intra-tumour heterogeneity mapping for a new patient). Because H&E whole-slide images are 2D and high resolution, they were spatially tiled into image patches, and patch-level embeddings were extracted using CONCH, a domain-specific pathology foundation model [8].

A 3D convolutional model was used to convert the PET tumour uptake to a latent representation, which was then used to identify sub-parts of the tumour most related to the tissue-level phenotype information learned from tiled H&E derived embeddings. The teacher network was first trained and implicit tumour-volume locations were treated as receptive field derived-voxels (iTS-HVox) and softly associated with H&E-derived phenotype prototypes. The teacher–student distillation step was used to train the student network to reproduce the teacher's PET embeddings and hotspot representations. The code will be open sourced at GitHub.

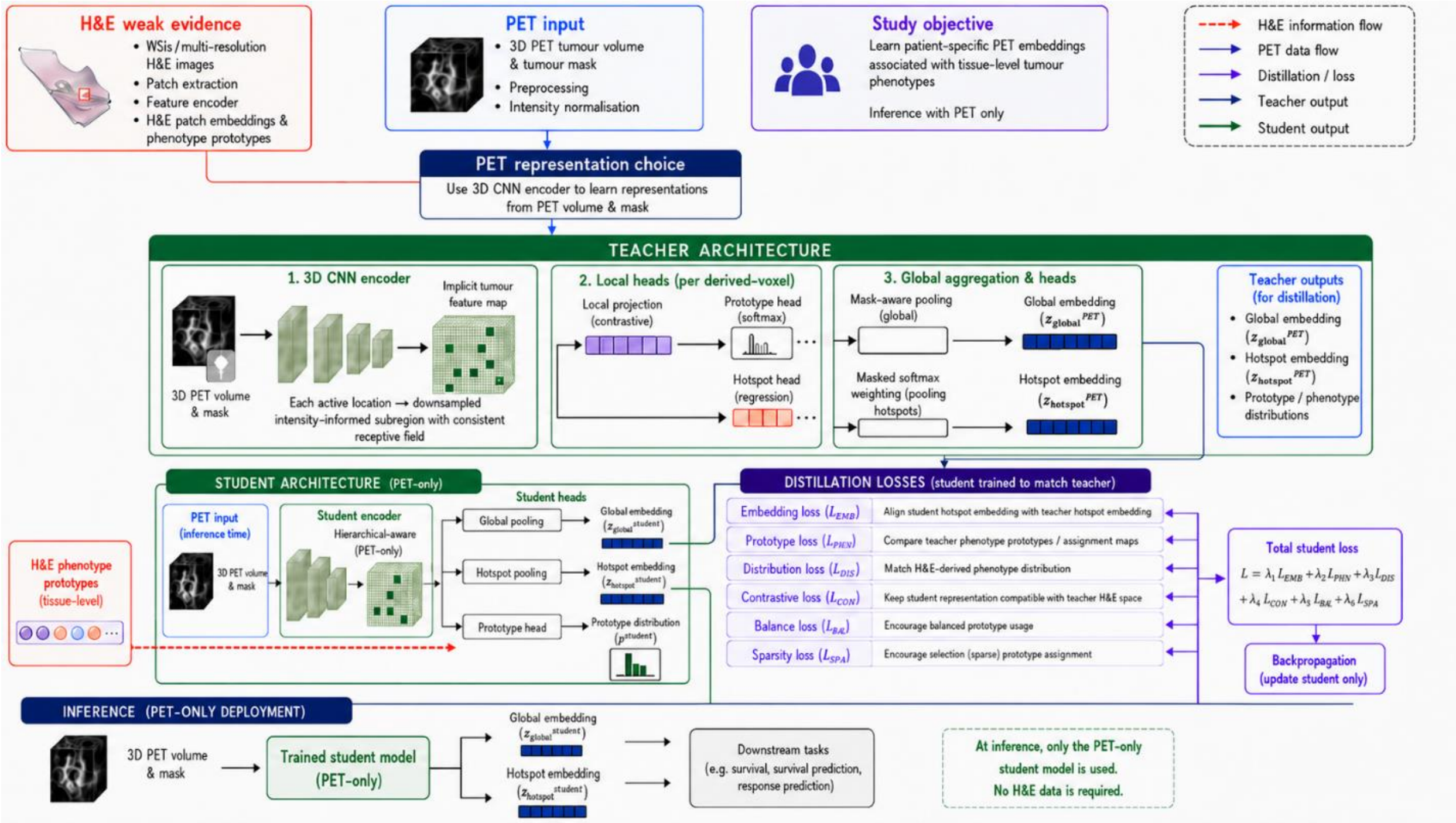

Figure 1. The schematic shows how H&E-derived weak supervision was used during teacher training to learn pathology-informed PET representations, while preserving PET-only inference through student distillation. During training the model processed the full PET tumour hotspot to learn internal pseudo-voxel representations linked to H&E-derived phenotype prototypes. Teacher outputs, including global PET embeddings, hotspot-conditioned embeddings and phenotype/prototype structure, were distilled into a PET-only student model. At inference, only the student model was used for downstream retrieval and virtual-biopsy analysis.

2.2.1. Patient cohort and multimodal data

We used multimodal data from 110 patients with oesophageal cancer retrospectively recruited from the Princess Alexandra Hospital, Brisbane, Australia, under institutional ethics approval 2022/HE002261. The dataset included PET imaging, whole-slide histopathology images and bulk transcriptomic data. For the PET component, PET images were converted to NIfTI format, and the primary tumour region was manually segmented using ITK-SNAP by experienced clinicians. In brief tumour-containing PET slices were identified from the corresponding clinician-annotated segmentation masks, which delineated the primary tumour region within the oesophageal/gastro-oesophageal area. For the histopathology component, tissue samples obtained at endoscopy were processed as whole-slide images. Within each WSI, pathologists annotated regions corresponding to normal tissue, tumour tissue and tumour differentiation patterns. These annotated regions were subsequently tiled and used as histopathology inputs. For the genomics component, matched endoscopic tissue samples were processed for bulk RNA extraction using the Qiagen AllPrep DNA/RNA Mini Kit according to the manufacturer's protocol. RNA sequencing libraries were prepared using the TruSeq Stranded mRNA Library Prep Kit and sequenced on an Illumina NovaSeq X platform using 150 bp paired-end sequencing.

The derivation of tile-level histopathology classes from WSI and bulk RNA data has been described in our previous Agent-SPI-WSI work [15]. In the present study, these histopathology-derived tile-level phenotype classes were used together with PET imaging to construct patient-level multimodal pairs for model training. For evaluation we have used two clinical variables, i.e. clinical stages defined using the PET images and tumor differentiation defined using the histopathology images. Model performance was assessed using five-fold patient-level cross-validation. Within each fold, patients were split 70:10:20 train: validation: test distribution. The validation subset was used for model selection and hyperparameter monitoring, while the held-out test subset was used for fold-wise performance evaluation.

2.3. Three teacher loss functions

The following steps were needed prior to computing individual losses: i) H&E whole-slide images were divided into patches, ii) each H&E patch was embedded using CONCH, iii) H&E patch embeddings were clustered to assign a patient-level tissue phenotype label (as performed in our previous work; Agent SPI-WSI [15], iv) PET tumour volumes were implicitly divided into 3D derived voxel tiles, and v) PET embeddings were aggregated into a patient-level PET representation.

2.3.1. Patient-level H&E phenotype loss ($\mathcal{L}_{H\&E}$)

We let $p$ define a patient from a total of $P$ patients with $i$ corresponding to a specific PET tile from a total of $N_p$ for that patient, then the global patient-level PET embedding ($\mathbf{z}_p^{global}$) obtained by pooling PET tile features was defined as:

$$\mathbf{z}_p^{global} = \text{Pool}[\mathbf{h}_{p,i}]_{i=1}^{N_p}$$

where the vector $\mathbf{h}_{p,i}$ of length $R^{256}$ denotes the feature representation of PET tile $i$ for patient $p$, and the Pool$(\cdot)$ operation was the aggregation function over PET tiles (mean or attention). If we consider $\hat{\mathbf{y}}_p$ as the predicted probability distribution over H&E-derived phenotype classes for patient $p$ and $\hat{y}_{p,c}$ as the predicted probability for class $c$, then the patient level phenotype loss ($\mathcal{L}_{H\&E}$) measures how well the overall PET tumor representation can predict the patient-level tissue phenotype derived from H&E as:

$$\mathcal{L}_{H\&E} = -\frac{1}{P}\sum_{p=1}^{P}\sum_{c=1}^{C} y_{p,c}\log(\hat{y}_{p,c}), \qquad \hat{\mathbf{y}}_p = \text{softmax}(\mathbf{W}_s\mathbf{z}_p^{global} + \mathbf{b}_\mathbf{s}),$$

where $\mathbf{W}_s$ and $\mathbf{b}_s$ are the learnable weight matrix and bias of the patient-level phenotype prediction head with $C$ the number of phenotype classes.

2.3.2. PET hotspot consistency loss ($\mathcal{L}_{HOT}$)

Assuming $T_{p,i}$ denotes the spatial location of PET tile $i$ and $M_p$ the tumour mask, then the PET-derived normalised hotspot target ($r_{p,i}$) was defined as:

$$\tilde{r}_{p,i} = \frac{r_{p,i}}{\sum_{j=1}^{N_p} r_{p,j} + \epsilon}, \qquad r_{p,i} = \frac{1}{|T_{p,i} \cap M_p|} \sum_{v \in T_{p,i} \cap M_p} \mathbf{1}\left[I_p(v) \geq \tau_p\right],$$

where $T_{p,i} \cap M_p$ captures tumour voxels only contained within the PET tile with $v$ denoting a PET voxel location and $I_p(v)$ the corresponding PET voxel intensity. The threshold $\tau_p$ was set to limit to high-uptake voxels and was patient specific., and the indicator function $\mathbf{1}[\cdot]$ returns 1 when the condition is true and 0 otherwise. The value for $\epsilon$ was chosen arbitrarily small for numerical stability. By defining $a_{p,i}$ as the raw hotspot or attention score predicted for a PET tile and $\alpha_{p,i}$ the corresponding normalised attention weight, the PET hotspot consistency loss was computed as:

$$\mathcal{L}_{HOT} = -\frac{1}{P} \sum_{p=1}^{P} \sum_{i=1}^{N_p} \tilde{r}_{p,i} \log(\alpha_{p,i}), \qquad \alpha_{p,i} = \frac{\exp(a_{p,i})}{\sum_{j=1}^{N_p} \exp(a_{p,j})},$$

2.3.3. Derived-voxel phenotype supervision ($\mathcal{L}_{OT}$)

We considered $\mathbf{w}_{p,v}$ as the feature vector of PET derived voxel $v$ from patient $p$, where a derived voxel was strictly within the internal tumour feature map produced by the convolution encoder, and let the vector $\mathbf{q}_k^{HE}$ denote H&E-derived phenotype prototype $k$ from $K$ number of them. For each patient $p$, we constructed a similarity matrix between PET-derived voxel features and H&E phenotype prototypes. The entropic transport plan $\Pi_{v,k}^{p}$was then defined as a non-negative assignment matrix that softly transports PET-derived voxel $v$ to H&E phenotype prototype $k$. Given matrix $\Pi_{v,k}$ as the entropic transport plan for a patient $p$ then:

$$\Pi_p^* = \arg\min_{\Pi \in U(a,b)} \sum_{i=1}^{M_p} \sum_{k=1}^{K} \Pi_{v,k}\, C_{p,v,k} + \beta \sum_{i=1}^{M_p} \sum_{k=1}^{K} \Pi_{v,k} \left(\log \Pi_{v,k} - 1\right),$$

was the optimal transport (OT) plan with $C_{p,v,k} = 1 - \mathrm{sim}(\mathbf{w}_{p,v}, \mathbf{q}_k^{HE})$ and $\mathrm{sim}(\cdot)$ as the cosine similarity, and the set $U(a,b)$ contained transport matrices with row and column marginals $a$ and $b$, where $a$ described the distribution of mass over PET pseudo-voxels and $b$ the distribution of mass over H&E phenotype prototypes. The matrix entry $\Pi_{p,v,k}^*$ was therefore the soft assignment strength between derived voxel $v$ and prototype $k$. If $\beta$ was set as a positive number to control the smoothness of the transport plan, then:

$$\tilde{y}_{p,v,k}^{OT} = \frac{\Pi_{p,v,k}^*}{\sum_{l=1}^{K} \Pi_{p,v,l}^* + \epsilon}, \qquad \hat{y}_{p,v}^{proto} = \mathrm{softmax}(\mathbf{W}_p v_{p,v} + \mathbf{b}_p),$$

where $\tilde{y}_{p,i,k}^{OT}$ could be considered as the OT-derived soft pseudo-label for derived-voxel $v$ and prototype $k$, and $\hat{y}_{p,v}^{proto}$ was the model-predicted prototype distribution. The vectors $\mathbf{W}_p$ and $\mathbf{b}_p$ were the derived-voxel prediction head weights and bias. The optimal transport loss softly assigned PET derived-voxels to H&E-derived phenotype prototypes as:

$$\mathcal{L}_{OT} = -\frac{1}{\sum_p M_p} \sum_{p=1}^{P} \sum_{i=1}^{M_p} \sum_{k=1}^{K} \tilde{y}_{p,v,k}^{OT} \log(\hat{y}_{p,v,k}^{proto}).$$

2.4. Teacher architecture and training

Each patient's segmented PET tumour was resized to a fixed 3D input shape, $X_p \in R^{1\times64\times64\times64}$ before being passed to iTS-HVox encoder. The encoder consisted of repeated convolutional blocks composed of 3D convolution, normalisation and non-linear activation. Three strided 3D convolutions block with stride was used to progressively downsample the input and generate a compact internal 3D tumour feature map, $\mathbf{F}_p \in \mathbb{R}^{256\times8\times8\times8}$, where $8\times8\times8$ is the downsampled internal spatial grid produced by the strided 3D convolutional encoder. Each active spatial location $v$ in this feature map was treated as a model-derived voxel and represented by a 256-dimensional vector $\mathbf{w}_{p,v} \in \mathbb{R}^{256}$. A model-derived voxel is therefore not a raw PET voxel, nor a manually cropped anatomical cube, but an internal learned representation of a PET tumour subregion produced by the encoder. Active model-derived PET voxels were were used for subsequent hotspot weighting, prototype assignment and patient-level embedding aggregation.

The iTS-HVox teacher included local projection head mapping of each derived-voxel feature into a lower-dimensional embedding space. The local projection head was implemented using $1\times1\times1$ convolutional layers, optionally followed by normalisation and non-linear activation, to project each local feature into a 128-dimensional embedding space, $\mathbf{g}_{p,v} \in \mathbb{R}^{128}$ . Because a 1×1×1 convolution was an independent spatial operation, it transformed the feature vector of each derived voxel without mixing information from neighbours. This allowed each tumour subregion to obtain its own local embedding while preserving the spatial organisation of the internal PET feature map.

The projected derived-voxel embedding was then passed to a prototype prediction head. This head estimated how strongly each derived voxel was associated with a set of H&E-derived phenotype prototypes. These prototypes represented tissue-level phenotype patterns learned from H&E information during training. The prototype head produced a probability distribution over phenotype prototypes for each derived voxel, typically using a convolutional layer followed by softmax normalisation. Thus, the model estimated whether its PET feature pattern was more consistent with one phenotype prototype or another at each derived voxel. In parallel, H&E patch embeddings were summarised into tissue-level phenotype prototypes and projected into the same embedding space using an H&E/prototype projection head.

Teacher architecture further used local derived-voxel head and global patient-level projection head. Local derived-voxel head estimated the relative scores (ranking) of each derived voxel. These scores were normalised using a masked softmax operation over the active tumour locations, and to form a hotspot-conditioned PET embedding through weighted pooling. A separate global PET embedding was obtained by mask-aware pooling of the internal tumour feature map.

During training, projection head placed PET-derived voxels and H&E phenotype prototypes into the same representational language. Optimal transport loss function then chose how strongly each PET-derived voxel should be associated with each H&E-derived phenotype prototype. Here, each PET-derived voxel and each H&E-derived phenotype prototype were compared using cosine similarity. A cost matrix was then constructed, where each entry measured how dissimilar a PET-derived voxel was from an H&E-derived phenotype prototype. Entropic optimal transport was applied to this cost matrix to obtain a soft assignment between PET-derived voxels and H&E phenotype prototypes. For iTS-HVox, the combined teacher loss was:

$$\mathcal{L}_{iTS-HVox} = \mathcal{L}_{H\&E} + \lambda_{HOT}\mathcal{L}_{HOT} + \lambda_{OT}\mathcal{L}_{OT},$$

with $\mathcal{L}_{H\&E}$ as the patient-level H&E phenotype loss, $\mathcal{L}_{HOT}$ the PET hotspot consistency loss, $\mathcal{L}_{OT}$ the entropic optimal-transport-based derived voxel phenotype supervision loss, and scalars $\lambda_{HOT}$ and $\lambda_{OT}$ controlled the contribution from the hotspot and OT-based pseudo-voxel supervision terms.

2.5. Teacher–student distillation for PET-only inference

2.5.1. Student Architecture

The student architecture mirrored that of the teacher without all H&E-dependent components. The student network was the PET-only inference model trained after completion of teacher training. Its role was to reproduce the teacher-derived PET representations using PET input alone. For the implicit iTS-HVox setting, the student followed the PET branch of the iTS-HVox teacher.

2.5.2. Distillation loss functions

Embedding loss: The embedding distillation term aligned the student hotspot embedding with the teacher representation using:

$$\mathcal{L}_{EMB} = \frac{1}{P}\sum_{p=1}^{P}\left\|\tilde{\mathbf{z}}_{p,S}^{hot} - \mathrm{sg}\left(\tilde{\mathbf{z}}_{p,T}^{fuse}\right)\right\|_2^2,$$

where $\tilde{\mathbf{z}}_{p,S}^{hot}$ is the normalised hotspot-conditioned PET embedding vector generated by the student (*S*) for patient $p$, and $\tilde{\mathbf{z}}_{p,T}^{fuse}$ is the normalised fused teacher (*T*) representation, which contains PET related information guided by H&E-derived weak supervision. The $\mathrm{sg}(\cdot)$ operator denotes the stop-gradient, meaning that the teacher target is fixed during the training of the student network.

Phenotype loss: The prototype preservation term encouraged the student to reproduce the teacher phenotype-prototype assignment pattern via:

$$\mathcal{L}_{PHE} = \frac{1}{\sum_p V_p}\sum_{p=1}^{P}\sum_{v=1}^{V_p}\mathrm{KL}\left(\mathbf{q}_{p,v,T} \parallel \mathbf{q}_{p,v,S}\right),$$

where $v$ indexes a local PET unit; in the explicit PET-tile models, a local unit corresponds to a PET tile and in B3iTS-HVox it corresponds to a PET-derived pseudo-voxel. The integer $V_p$ is the number of local PET units for a patient, $\mathbf{q}_{p,v,T}$ is the teacher phenotype-prototype probability distribution vector for local unit $v$, and $\mathbf{q}_{p,v,S}$ is the corresponding student version, with the Kullback-Leibler divergence denoted by $\mathrm{KL}(\cdot\|\cdot)$. This term transfers the teacher prototype structure to the PET-only student model.

Distribution loss: The term was used to encourage the student to retain the teacher's H&E-informed patient-level phenotype organisation according to:

$$\mathcal{L}_{DIS} = \frac{1}{P}\sum_{p=1}^{P} \mathrm{KL}\left(\mathbf{d}_{p,T} \parallel \mathbf{d}_{p,S}\right),$$

where $\mathbf{d}_{p,T}$ captures the teacher and $\mathbf{d}_{p,S}$ the corresponding student phenotype distribution. These distributions were obtained by aggregating local prototype probabilities across tumour units.

OT loss: This loss is a cross-entropy term between teacher-derived OT assignments and student prototype probabilities. The student optimal-transport consistency term encouraged the student local-unit assignments to follow teacher-derived soft OT targets via:

$$\mathcal{L}_{OT}^{S} = -\frac{1}{\sum_p V_p}\sum_{p=1}^{P}\sum_{v=1}^{V_p}\sum_{k=1}^{K} \tilde{y}_{p,v,k}^{OT,T}\log\left(q_{p,v,k,S}\right),$$

where $\tilde{y}_{p,v,k}^{OT,T}$ is the teacher-derived OT soft target assigning local PET unit $v$ from patient $p$ and prototype $k$. The vector element $q_{p,v,k,S}$ is the student-predicted probability at the same local unit.

Contrastive loss: This loss pulls the student PET representation towards the matched teacher H&E-informed representation and separates it from unmatched patient representations according to:

$$\mathcal{L}_{CON} = -\frac{1}{P}\sum_{p=1}^{P}\log\frac{\exp\left(\frac{\mathrm{sim}\left(\tilde{\mathbf{z}}_{p,S}^{hot}, \tilde{\mathbf{h}}_{p,T}^{HE}\right)}{\tau_{CON}}\right)}{\sum_{j=1}^{P}\exp\left(\frac{\mathrm{sim}\left(\tilde{\mathbf{z}}_{p,S}^{hot}, \tilde{\mathbf{h}}_{j,T}^{HE}\right)}{\tau_{CON}}\right)},$$

with $\tilde{\mathbf{z}}_{p,S}^{hot}$ the student hotspot embedding and $\tilde{\mathbf{h}}_{p,T}^{HE}$ teacher H&E-informed representation. The function $\mathrm{sim}(\cdot)$ denotes the cosine similarity and $\tau_{CON}$ is a temperature parameter.

Balanced loss: This loss was used to discourage the student from assigning most tumour units to only one or a few prototypes by:

$$\mathcal{L}_{BAL} = \sum_{k=1}^{K}\bar{q}_{k,S}\log\left(\frac{\bar{q}_{k,S}}{1/K}\right), \qquad \bar{q}_{k,S} = \frac{1}{\sum_p V_p}\sum_{p=1}^{P}\sum_{v=1}^{V_p} q_{p,v,k,S},$$

where $\bar{q}_{k,S}$ is the average student usage of prototype $k$ across all local PET units in the batch. The term $1/K$ represents uniform prototype usage with the $K = 3$, i.e. H&E-informed phenotype prototypes.

Sparsity loss: The sparsity term encouraged selective local prototype assignment according to:

$$\mathcal{L}_{SPA} = -\frac{1}{\sum_p V_p}\sum_{p=1}^{P}\sum_{v=1}^{V_p}\sum_{k=1}^{K} q_{p,v,k,S}\log(q_{p,v,k,S}).$$

Minimising of this term encourages each PET tile or pseudo-voxel to assign high probability to a smaller number of relevant prototypes rather than distributing probability uniformly across all prototypes.

2.5.3. Student training using weighted distillation losses ($\mathcal{L}_S$)

The student objective combined the weighted combination of embedding distillation (*e*), prototype preservation, distribution matching, student optimal-transport consistency, contrastive compatibility and prototype regularisation:

$$\mathcal{L}_S = \lambda_{EMB}\mathcal{L}_{EMB} + \lambda_{PRE}\mathcal{L}_{PRE} + \lambda_{DIS}\mathcal{L}_{DIS} + \lambda_{OT}\mathcal{L}_{OT}^{S} + \lambda_{CON}\mathcal{L}_{CON} + \lambda_{BAL}\mathcal{L}_{BAL} + \lambda_{SPA}\mathcal{L}_{SPA},$$

where scalars $\lambda_{EMB}$, $\lambda_{PRE}$, $\lambda_{DIS}$, $\lambda_{OT}$, $\lambda_{CON}$, $\lambda_{BAL}$ and $\lambda_{SPA}$ control the relative contribution of each loss term with hyperparameters weights 1.0, 1.0, 0.5, 1.0, 0.2, 0.03 and 0.01 respectively. The hyperparameters were selected based on trial and error during experiment design.

2.6. Content-based medical image retrieval

Learned PET representation was evaluated using application in content-based medical image retrieval [16-20]. Retrieval relevance was defined with respect to clinical staging than tumor differentiation. A retrieved patient was considered relevant if that patient shared the same endpoint label as the query patient. Retrieval performance was evaluated at using mean average precision ($k = 3, 5, 7$), normalised discounted cumulative gain, and mean reciprocal rank [19, 20].

Mean average precision at $k$: Mean average precision at $\boldsymbol{k}$ ($\boldsymbol{mAP@k}$) was used as the primary retrieval metric because it measures both the presence of relevant retrieved patients and their ordering in the ranked list. For query patient $f$, precision at rank $i$ is defined as**:**

$$P_f(i) = \frac{1}{i}\sum_{j=1}^{i} \mathrm{rel}_f(j),$$

where $\mathrm{rel}_q(j) = 1$ if the patient retrieved at rank $j$ is relevant to query $f$, and $\mathrm{rel}_f(j) = 0$, otherwise. Average precision at $k$ was calculated using:

$$AP_f@k = \frac{1}{\min(R_f, k)}\sum_{i=1}^{k} P_f(i)\mathrm{rel}_f(i),$$

where $R_f$ denotes the total number of relevant patients available for query $f$. This then leads to the mean average precision across all query patients:

$$mAP@k = \frac{1}{F}\sum_{f=1}^{F} AP_f@k.$$

Normalised discounted cumulative gain at $k$: Normalised discounted cumulative gain at $\boldsymbol{k}$ ($\boldsymbol{nDCG@k}$) evaluates ranking quality while giving greater weight to relevant patients that appear near the top of the retrieval list. The discounted cumulative gain was:

$$DCG_f@k = \sum_{i=1}^{k} \frac{\mathrm{rel}_f(i)}{\log_2(i+1)}.$$

The ideal discounted cumulative gain, $IDCG_f@k$, was obtained by sorting the retrieval list so that all relevant patients appeared first. The normalised score was calculated according to:

$$nDCG_f@k = \frac{DCG_f@k}{IDCG_f@k}, \qquad nDCG@k = \frac{1}{F}\sum_{f=1}^{F} nDCG_f@k,$$

with $nDCG@k$ defining the measure across all patients.

Mean reciprocal rank at $k$**:** Mean reciprocal rank at $\boldsymbol{k}$ ($\boldsymbol{MRR@k}$) is used to measure how early the first relevant neighbour appears in the ranked retrieval list. For query patient $f$, let $r_f$ be the rank of the first relevant retrieved patient within the top $k$. The reciprocal rank is then**:**

$$RR_f@k = \begin{cases} \frac{1}{r_f}, & \text{if a relevant patient is retrieved within the top } k, \\ 0, & \text{otherwise,} \end{cases}$$

with the mean as:

$$MRR@k = \frac{1}{F}\sum_{f=1}^{F} RR_f@k.$$

2.6.1 Benchmarking and study comparisons

For every fold, the training patients formed the retrieval gallery and held-out test patients formed the query set. The benchmarking analysis was designed to evaluate whether histopathology-informed PET hotspot representations improved content-based medical image retrieval compared with global, weakly supervised, conventional radiomics PET-derived [20, 21] and two previous CNN based methods using the above metrices. In the radiomics baseline, the segmented PET tumour volume was used to extract quantitative features describing tumour intensity, shape and texture, thereby representing each patient using a fixed feature vector derived from the original PET signal rather than learned embeddings. These features were then used for downstream similarity/retrieval analysis. In the CNN baseline, each 3D PET tumour volume was converted into maximum-intensity-projection views [22], providing a compact 2D representation of tumour uptake distribution, and a convolutional neural network was trained to learn image-level PET embeddings from these projections [23].

2.7. Three ablation experiments

We conducted ablation experiments to evaluate the PET-specific training representations. In the first experiment, to test the efficiency of the implicit tilling, the 3D PET tumour volume was explicitly divided into fixed-size 3D tiles. Each patient was represented by a set of PET tile embeddings where $\mathbf{x}_{p,i}^{PET} \in \mathbb{R}^{d_{PET}}$ was a precomputed embedding for the $i$-th PET tile, and $\mathbf{c}_{p,i} \in \mathbb{Z}^3$ was its tile-centre coordinate in voxel space. In embedding mode, $d_{PET} = 512$ and PET embeddings are produced by training a separate 3D convolutional autoencoder (note, in Figure 1 Module replaced by an explicitly trained autoencoder). The H&E-derived embeddings were subsequently used outside the PET embedding encoding model to provide weak tissue-level phenotype supervision.

In the second experiment, we were interested in moving from weak patient-level supervision toward progressively stronger cross-modal supervision. As such, three different types of training strategies were used to evaluate the effect of loss components under the explicit tilling:

S1) Baseline strategy - apply patient-level H&E-derived labels subtype supervision and PET hotspot consistency loss to test whether patient-level H&E phenotype supervision and PET hotspot consistency alone were sufficient to shape useful PET representations;

S2) Baseline plus raw H&E-to-PET tile pseudo-label transfer - test whether direct nearest-neighbour cross-modal matching was sufficient with the purpose of understanding whether direct raw-space H&E-to-PET pseudo-label transfer could improve PET hotspot representation learning beyond the S1 baseline;

S3) Addition of supervised PET–H&E latent alignment before pseudo-label transfer - test whether cross-modal alignment improved the reliability of H&E-derived supervision.

These methodological ablations were designed to isolate the contribution of each supervision strategy. An improvement from S1 to S2 would suggest that raw pathology-derived pseudo-label transfer provides additional information beyond patient-level supervision. Further, improvement from S2 to S3 would indicate that PET–H&E latent alignment improves the reliability of cross-modal pseudo-label transfer. Student used the same architecture as of teacher under each strategy.

The third experiment was designed to test whether the identified hotspots in PET tumour were functionally important for retrieval. Hotspot-conditioned embeddings were thought to emphasise tumour regions that contribute most strongly to patient-level similarity. Therefore, a faithful hotspot map should identify PET tumour regions whose removal meaningfully disrupts the retrieval representation. Conversely, if removing high-scoring attention regions has the same effect as removing random or low-scoring regions, the hotspot scores would be less likely to reflect functionally important PET evidence.

2.7.1. Explicitly trained autoencoder: Each PET tile was represented as a single-channel $5 \times 5 \times 5$ voxel tensor and passed through 3D convolutional encoder blocks. Each block consisted of a 3×3×3 convolution, normalisation and non-linear activation. Because the PET tile was spatially small, convolutions were applied with stride 1 and padding to preserve local volumetric structure rather than aggressively downsampling the input. The final encoder feature tensor was flattened and projected into a 512-dimensional latent representation. The decoder mirrored the encoder by first expanding the latent vector back to a compact 3D feature tensor and then applying reconstruction blocks followed by a final single-channel 3×3×3 convolution to reconstruct the original PET tile.

2.7.2. Raw H&E-to-PET tile pseudo-label loss ($\mathcal{L}_t^{raw}$)

This loss was used to test whether direct nearest-neighbour matching between raw PET and H&E-derived features is sufficient for cross-modal pseudo-label transfer. If $\mathbf{e}_{p,i}^{PET}$ is the raw explicitly trained embedding of PET tile $i$ from patient $p$, and $\mathbf{h}_{p,m}^{HE}$ is the raw embedding of H&E patch $m$ from the same patient. The index $m^*(i)$ defines the H&E patch most similar to PET tile $i$ in the raw embedding space based on cosine similarity (note, here we were forcing the embedding space similarity). The label $y_{p,m^*(i)}^{HE}$ was the H&E-derived phenotype label of the matched H&E patch. The term $\tilde{y}_{p,i}^{raw}$ was the weak pseudo-label assigned to PET tile $i$, and $\tilde{y}_{p,i,c}^{raw}$ was its value for phenotype class $c$. The prediction $\hat{y}_{p,i}^{tile}$ was the teacher's tile-level phenotype probability distribution for PET tile $i$, and $\hat{y}_{p,i,c}^{tile}$ was the predicted probability for class $c$. Then:

$$\mathcal{L}_t^{raw} = -\frac{1}{\sum_p N_p} \sum_{p=1}^{P} \sum_{i=1}^{N_p} \sum_{c=1}^{C} \tilde{y}_{p,i,c}^{raw} \log\left(\hat{y}_{p,i,c}^{tile}\right),$$

based on:

$$\tilde{\mathbf{y}}_{p,i}^{raw} = \mathbf{y}_{p,m^*(i)}^{HE}, \quad \hat{\mathbf{y}}_{p,i}^{tile} = \text{softmax}\left(\mathbf{W}_t \mathbf{h}_{p,i} + \mathbf{b}_t\right), \quad m^*(i) = \underset{m}{\text{argmax}} \text{sim}\left(\mathbf{e}_{p,i}^{PET}, \mathbf{h}_{p,m}^{HE}\right).$$

. matching was performed in the raw feature space and did not force PET-H&E spatial correspondence.

2.7.3. PET-H&E latent alignment loss ($\mathcal{L}_c$)

For $\mathbf{z}_p^{PET}$ and $\mathbf{z}_p^{HE}$ defined as patient-level PET and H&E representations for patient $p$, the functions $g_{PET}(\cdot)$ and $g_{HE}(\cdot)$ denote the projection networks that mapped PET and H&E representations into a shared latent space:

$$s_{p,l} = \frac{\left(\mathbf{u}_p^{PET}\right)^T \mathbf{u}_l^{H\&E}}{\tau_{CON}}, \quad \mathbf{u}_p^{PET} = \frac{g_{PET}\left(\mathbf{z}_p^{PET}\right)}{\left\|g_{PET}\left(\mathbf{z}_p^{PET}\right)\right\|_2}, \quad \mathbf{u}_p^{H\&E} = \frac{g_{H\&E}\left(\mathbf{z}_p^{H\&E}\right)}{\left\|g_{H\&E}\left(\mathbf{z}_p^{H\&E}\right)\right\|_2},$$

and from the $s_{p,l}$ score we computed the PET-H&E latent alignment loss:

$$\mathcal{L}_c = \frac{1}{2}\left(\mathcal{L}_c^{PET \to H\&E} + \mathcal{L}_c^{H\&E \to PET}\right),$$

$$\mathcal{L}_c^{PET \to H\&E} = -\frac{1}{B} \sum_{p=1}^{B} \log \frac{\exp\left(s_{p,p}\right)}{\sum_{l=1}^{B} \exp\left(s_{p,l}\right)}, \quad \mathcal{L}_c^{H\&E \to PET} = -\frac{1}{B} \sum_{p=1}^{B} \log \frac{\exp\left(s_{p,p}\right)}{\sum_{l=1}^{B} \exp\left(s_{l,p}\right)},$$

where $B$ defines the batch size. This loss was used to learn a shared latent space in which matched PET and H&E representations from the same patient were forced to be closer than unmatched representations from different patients.

2.7.4. Aligned H&E-to-PET tile pseudo-label loss ($\mathcal{L}_t^{align}$)

This loss was used to transfer H&E-derived pseudo-labels to PET tiles after PET and H&E were projected into the aligned latent space:

$$\mathcal{L}_t^{align} = -\frac{1}{\sum_p N_p} \sum_{p=1}^{P} \sum_{i=1}^{N_p} \sum_{c=1}^{C} \tilde{y}_{p,i,c}^{align} \log\left(\hat{y}_{p,i,c}^{tile}\right),$$

where

$$\tilde{\mathbf{y}}_{p,i}^{align} \stackrel{\text{def}}{=} \mathbf{y}_{p,m_{align}^{*}(i)}^{H\&E}, \quad m_{align}^{*}(i) = \underset{m}{\operatorname{argmax}} \operatorname{sim}\left(u_{p,i}^{PET}, u_{p,m}^{H\&E}\right),$$

and

$$\mathbf{u}_{p,i}^{PET} = \frac{g_{PET}\left(\mathbf{h}_{p,i}\right)}{\left\| g_{PET}\left(\mathbf{h}_{p,i}\right)\right\|_2}, \quad \mathbf{u}_{p,m}^{H\&E} = \frac{g_{H\&E}\left(\mathbf{e}_{p,m}^{H\&E}\right)}{\left\| g_{H\&E}\left(\mathbf{e}_{p,m}^{H\&E}\right)\right\|_2}.$$

The label $\mathbf{y}_{p,m_{align}^{*}(i)}^{H\&E}$ defined the H&E-derived phenotype label of the matched H&E patch, $\tilde{\mathbf{y}}_{p,i}^{align}$ was the weakly aligned pseudo-label and $\tilde{\mathbf{y}}_{p,i,c}^{align}$ the class-specific value. The prediction $\hat{y}_{p,i,c}^{tile}$ is the teacher's predicted tile-level probability for class $c$. Essentially, $\mathcal{L}_t^{align}$ conformed to the same classification form as $\mathcal{L}_t^{raw}$, but the matching was performed after PET-H&E latent alignment.

2.7.5. Architecture and training strategy

PET tumor volumes vary in size and shape implying each patient can have a different number of PET tiles to process. The teacher network included a projection layer to cater for different number of tiles and to produce consistent embeddings across patients. The projected tile features were aggregated using a learned tile-aggregation module, namely a learned-query attention block, to obtain a fixed-size patient-level PET representation.

Teacher network was followed by two separate branches producing complementary PET representations. The first was a global PET embedding, obtained by aggregating information across the tumour tile set. Here, a patient-level phenotype prediction head used the global PET embedding to predict the H&E-derived patient-level phenotype label. The head consisted of a linear classification layer followed by a softmax output.

The second branch was a hotspot-conditioned PET embedding ($\mathbf{z}_p^{hot}$), obtained by assigning attention or hotspot weights to PET tiles and then forming a weighted combination of tile features. The hotspot scoring head was implemented as a small prediction head operating on PET tile features and normalised across tiles to define the relative contribution of each tile to the hotspot representation:

$$\mathbf{z}_p^{hot} = \sum_{i=1}^{N_p} \alpha_{p,i} \, \mathbf{h}_{p,i},$$

where $\alpha_{p,i}$ is the normalised attention weight.

Ablation strategy S1): The teacher in Figure 1 used PET tile embeddings, PET-derived hotspot evidence and patient-level H&E-derived phenotype labels. For this baseline experiment, H&E contributed only as a weak patient-level supervisory label. The teacher was trained to predict the patient-level H&E-derived phenotype from the PET representation and to assign hotspot weights that were consistent with PET-derived high-uptake tumour evidence using:

$$\mathcal{L}_{\text{S1}} = \mathcal{L}_{PHE} + \lambda_{HOT}\mathcal{L}_{HOT}.$$

Ablation strategy S2): The teacher retained the same core architecture as in S1 but tile-level H&E-derived pseudo-label supervision was added. Individual PET tiles were additionally matched to H&E-derived representations using nearest-neighbour matching in the unaligned feature space. The matched H&E phenotype was treated as a weak pseudo-label for the corresponding PET tile using:

$$\mathcal{L}_{\text{S2}} = \mathcal{L}_{\text{S1}} + \lambda_t\mathcal{L}_t^{raw}.$$

Ablation strategy S3): The teacher training was change how the H&E-derived pseudo-labels were transferred to PET tiles. Instead of assigning pseudo-labels in the raw feature space, in S3 a PET–H&E latent alignment was first learnt. Patient-level PET and H&E representations were projected to a shared latent space using separate PET and H&E projection heads, each consisting of linear projection layers followed by normalisation. A symmetric contrastive alignment objective was then used to bring matched PET–H&E patient pairs closer together and push unmatched pairs apart. After this alignment step, H&E-derived pseudo-labels were transferred to PET tiles in the aligned latent space according to the loss:

$$\mathcal{L}_{\text{S3}} = \mathcal{L}_{\text{S1}} + \lambda_c\mathcal{L}_c + \lambda_t\mathcal{L}_t^{align}.$$

This alignment should be interpreted as a representation-level training strategy and does not imply spatial registration between PET and H&E.

2.7.6. Deletion of hotspots

In the third experiment for each query patient, PET regions were removed from the query representation, and retrieval was repeated against the unchanged training-fold. In the explicit-tile models, the removed units were PET tiles. In B3iTS-HVox, the removed units were PET derived voxels from the internal tumour feature map. The search training fold were kept unchanged so that any change in retrieval performance could be attributed specifically to perturbing the held-out query representation. Deletion fractions of 5%, 10%, 20% and 30% were considered.

Specifically, three deletion conditions were compared. High-attention deletion removed the PET tumour regions with the highest hotspot scores. Low-attention deletion removed the PET tumour regions with the lowest hotspot scores. Random deletion removed the same number of PET tumour regions selected at random and served as a matched control for information removed. After deletion, the query embedding was recomputed, and the same content-based retrieval evaluation was repeated.

### 2.8. Concordance with histology features and tumour volume

To test if voxel-derived hotspot prototype classes (aggressiveness, instability and suppression) show relationships with H&E tissue fractions and if these prototype classes differ in original PET uptake, we calculated PET uptake enrichment as the difference between the mean original PET intensity of derived-voxels assigned to that class and the mean original PET intensity across the active tumour region. A focused set of five histomic features was selected to provide an interpretable validation of PET-derived prototype classes against H&E morphology [24]. The selected features were chosen to capture complementary aspects of tumour aggressiveness: cellularity, nuclear size, nuclear shape irregularity, contour compactness and texture heterogeneity. This restricted feature set was preferred over using the full histomics feature space for initial validation because the aim was not to optimise a predictive histology model, but to test whether PET-derived hotspot prototype classes showed plausible patient-level concordance with interpretable H&E-derived morphology. We then correlated class-specific PET uptake enrichment against nuclear density, nuclear area, nuclear eccentricity, nuclear solidity and nuclear Haralick entropy. Correlations were computed across patients using Spearman's rank correlation

In addition, we tested whether clinical stage III and Stage IV tumours differ in global tumour burden and in the relative composition of ITS-HVox PET-derived hotspot classes. It is to evaluate the relative distribution of PET-derived tumour habitats differs between clinical stages. In each fold, for every patient, each voxel was assigned to one of the three prototype classes. The percentage of tumour volume assigned to each class was then calculated and compared against the total tumor volume. Statistical comparisons were performed using a two-sided Welch's independent-samples t-test.

## 3. Results

### 3.1. Retrieval Performance comparison

Figure 2 summarises the progression from PET-only hotspot learning to raw H&E transfer as an ablation, and then to aligned cross-modal and implicit pseudo-voxel supervision. For clinical-stage retrieval (A), iTS-HVox hotspot was the most consistent representation across *mAP@k*, *nDCG@k* and *MRR@k* at $k$ = 3, 5 and 7. It achieved the best *mAP@k* across all neighbourhood sizes, with approximate values of 0.56 at *mAP@*3, 0.51 at *mAP@*5 and 0.47 at *mAP@*7, outperforming PET radiomics, PET-MIP CNN baselines and ablations S1 to S3, which generally ranged from 0.40 to 0.53. iTS-HVox hotspot was also among the leading methods for *nDCG@k*, reaching approximately 0.62 at *nDCG@*3 and remaining competitive at *nDCG@*5–7, indicating that clinically stage-matched neighbours were ranked early in the retrieval list. The clearest advantage was observed for *MRR@k*, where iTS-HVox hotspot achieved approximately 0.74–0.78 across the $k$'s, showing that the first correct clinical-stage match was retrieved earlier than with competing methods. Compared with iTS-HVox global, the hotspot variant was generally superior, particularly for *mAP@k* and *MRR@k*, supporting selective pseudo-voxel hotspot pooling over global lesion-level PET aggregation. In contrast, explicit-tile hotspot/global differences were less consistent, and S2 did not show stable benefit, supporting the interpretation that raw H&E pseudo-label transfer alone is insufficient for robust stage-relevant retrieval.

Figure 2B further shows that tumour-differentiation retrieval, although performed using PET images at inference, remained informative for an H&E-defined phenotype. Across *mAP@k*, *nDCG@k* and *MRR@k* at $k$ = 3, 5 and 7, learned PET representations were broadly competitive with PET radiomics and PET-MIP CNN baselines, with B3iTS-HVox hotspot again showing the strongest and most stable early-rank performance. B3iTS-HVox hotspot achieved the highest or near-highest *mAP@k*, approximately 0.56 at *mAP@*3 and 0.50 at *mAP@*5, with performance remaining competitive at *mAP@*7 as method differences narrowed. For *nDCG@k*, iTS-HVox hotspot and iTS-HVox global remained among the leading methods, generally around 0.60–0.63 at lower $k$, indicating preserved ranking quality for tumour-differentiation-relevant neighbours. The strongest evidence was again observed for *MRR@k*, where B3iTS-HVox hotspot reached approximately 0.77–0.80 across $k$, suggesting that H&E-defined tumour-differentiation matches were retrieved very early using PET-only inference.

The S1-S3 ablation experiments were designed to test whether progressively more structured sources of supervision improve PET hotspot representation and retrieval. Based on Figure 2, S1 established that PET-only hotspot learning already provides a meaningful baseline, indicating that localised PET evidence contains clinically and phenotypically relevant retrieval signals. S2 did not consistently improve over S1, suggesting that raw H&E-derived pseudo-label transfer alone is insufficient and may introduce noisy supervision when PET and H&E embeddings are not aligned. The more competitive performance of S3 relative to S2 supports the need for cross-modal alignment before pathology-informed supervision can be transferred reliably. The strongest evidence came from iTS-HVox hotspot representation, which achieved the most consistent early-rank retrieval performance for clinical stage and tumour differentiation. These findings suggest that soft, spatially structured pseudo-voxel supervision provides additional benefit beyond hard tile-level pseudo-labels by allowing PET subregions to organise around phenotype-informed evidence without assuming direct PET–H&E spatial registration.

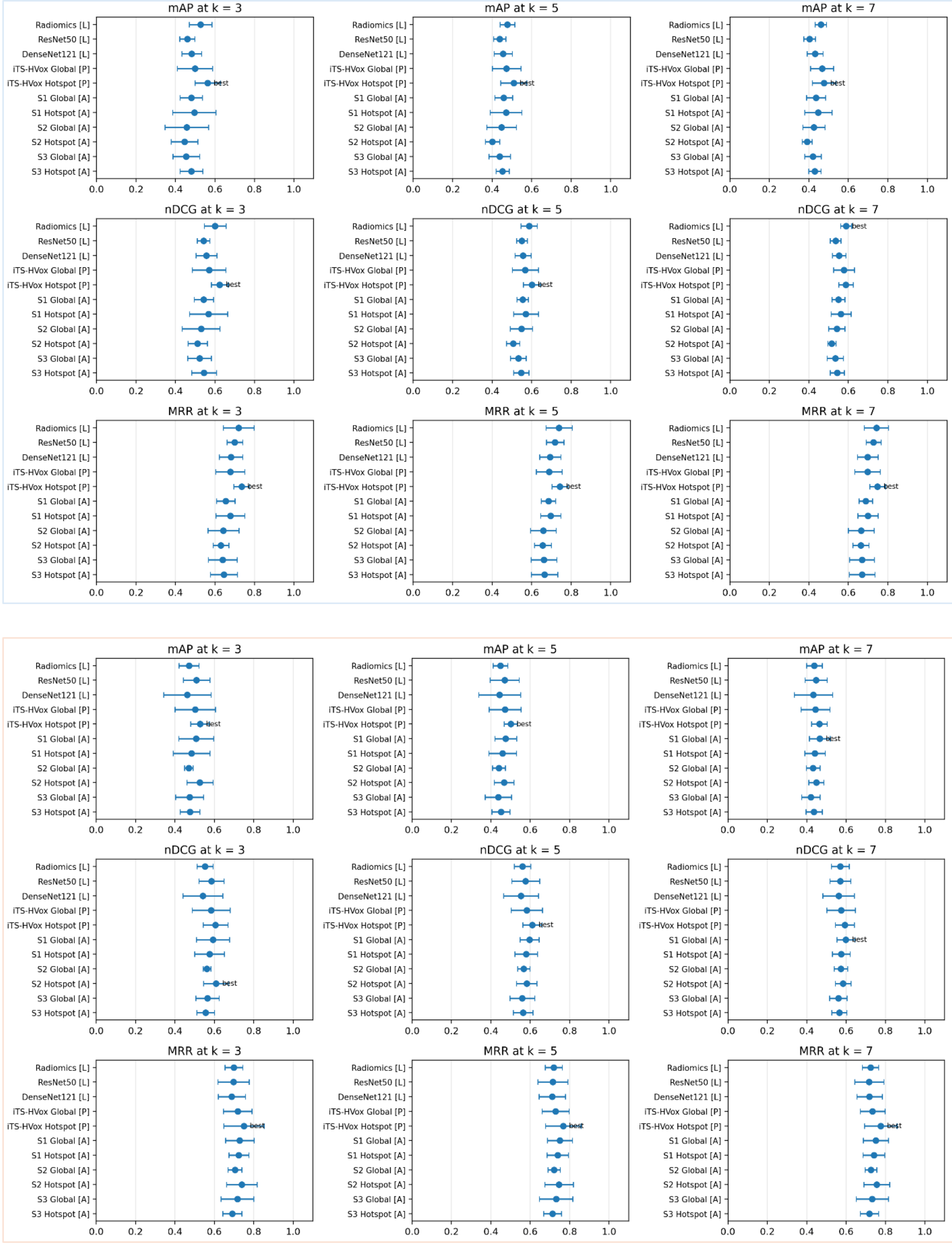

**Figure 2.** Clinical-stage (top) and tumor differentiation (bottom) retrieval performance across all folds. Rows show retrieval metrics, including *mAP@k*, *nDCG@k* and *MRR@k* while columns show increasing *k*. Methods included are conventional PET

radiomics, PET-MIP deep-feature Resnet50 and DenseNet121 according to literature [L], the proposed [P] iTS-HVox global and hotspot embeddings, and ablation [A] experiments. Bold points indicate the mean performance across cross-validation folds and error bars indicate standard deviation, and the best performance in each case has been labelled for visual interpretation.

### 3.2. Attention-faithfulness and sensitivity analysis

As shown in Figure 3, the deletion analysis highlights that the iTS-HVox hotspot representation is strongly dependent on a selective subset of high-evidence derived-voxel locations, rather than being driven by diffuse or random lesion-wide information. Figure 3A shows that top-attention deletion caused a monotonic increase in prototype-confidence drop, rising from approximately 0.004 to 0.018 from 5% to 30% deletion. Random deletion produced near-zero confidence changes, whereas low-attention deletion produced negative confidence changes, indicating that removing low-attention derived-voxels did not damage the dominant prototype signal and may even concentrate it. This pattern is consistent with a faithful hotspot mechanism, where high-attention pseudo-voxels carry the evidence most responsible for the learned phenotype-prototype assignment.

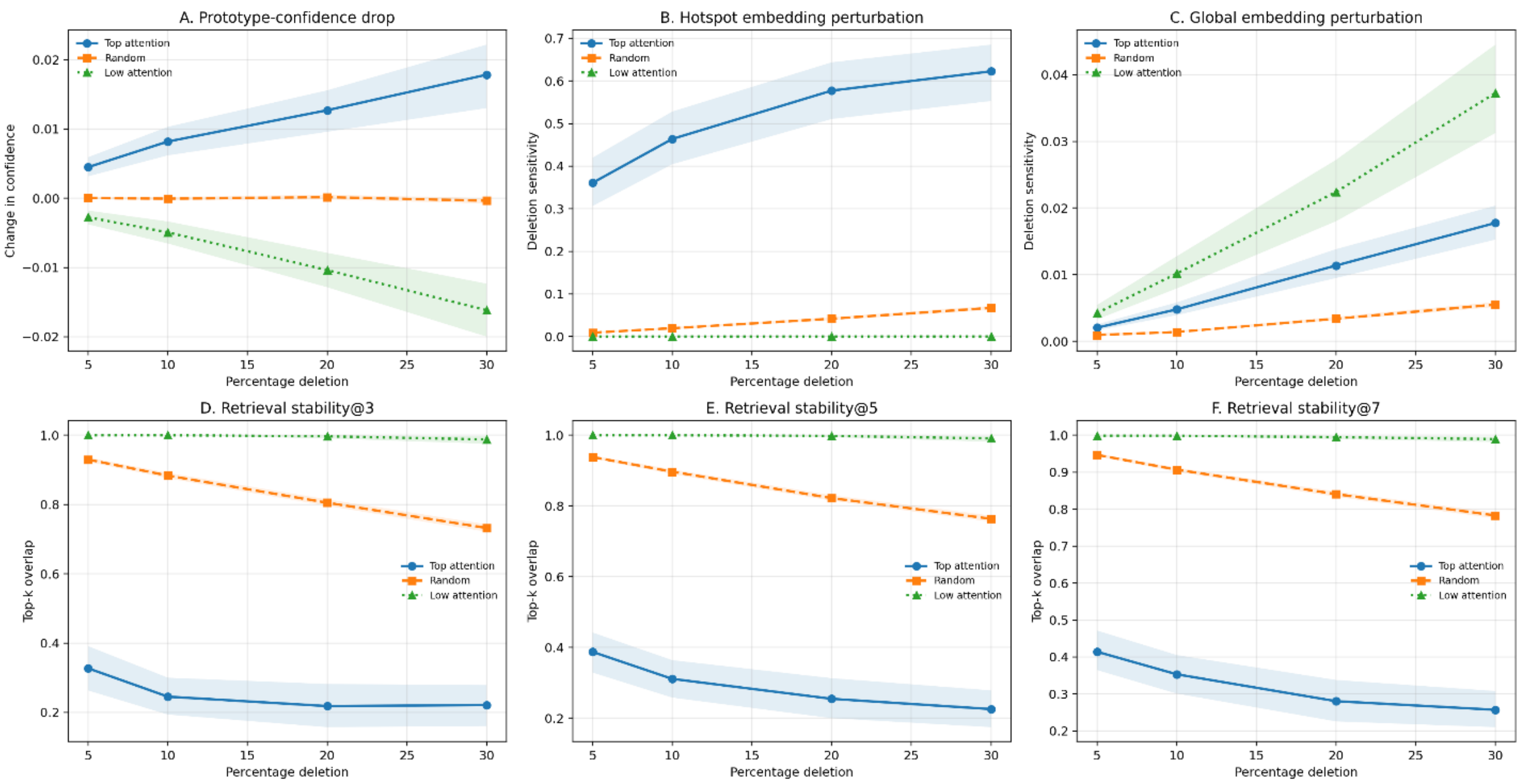

Figure 3**.** iTS-HVox pseudo-voxel deletion analysis for hotspot faithfulness. Attention faithfulness was evaluated by deleting top-attention, random or low-attention derived-voxels from held-out PET tumour representations at deletion fractions of 5%, 10%, 20% and 30%. (A–C) show the effect of deletion on prototype confidence, hotspot embedding perturbation and global embedding perturbation (1 − cosine (z_hot/global, z_hot/global) represents the deletion sensitivity), while (D–F) show retrieval-neighbour stability at $k = 3$, 5 and 7, measured as top-$k$ overlap before and after deletion.

In Figure 3B hotspot embedding showed the strongest deletion sensitivity. Removing top-attention pseudo-voxels increased hotspot-embedding perturbation from approximately 0.36 to 0.62 from 5% to 30% deletion. In contrast, random deletion produced only small perturbations, increasing from approximately 0.01 to 0.07 across the same range, and low-attention deletion produced almost no hotspot-embedding change. Taken together, the hotspot embedding is not equivalent to a global average representation; it is specifically dependent on high-attention pseudo-voxel evidence.

In Figure 3C global embedding perturbation showed that low-attention deletion perturbed the global embedding more strongly than top-attention deletion at higher deletion fractions. This is expected because the global embedding aggregates broader lesion-wide information, hence removal of many low-attention pseudo-voxels can still change the overall lesion average.

In Figure 3D-F retrieval-neighbour stability also decreased sharply after top-attention deletion. Stability @3, @5, @7 and @10 were all markedly lower for top-attention deletion than for random or low-attention deletion. At stability @5, for example, top-attention deletion reduced neighbour overlap from approximately 0.39 to 0.23 from 5% to 30% deletion, whereas random deletion retained high overlap and low-attention deletion remained close to 1.0. This demonstrates that the high-attention pseudo-voxels are not only important for the embedding vector itself but also for the downstream retrieval neighbourhood.

As shown in Figure 4A, the confidence-drop behaviour was less consistent. Ablation S1 showed a negative top-minus-random confidence effect, S2 remained close to zero or slightly negative, and S3 showed only a small positive effect. This suggests that explicit-tile attention is partly informative for retrieval representation. In Figure 4B, the explicit-tile models, the top-attention-minus-random analysis showed that deleting high-attention tiles generally perturbed the hotspot embedding more than random deletion, but the magnitude of this effect was modest. Across 5 to 30% deletion, S1, S2 and S3 showed small positive increases in hotspot-embedding disruption relative to random deletion, with S3 showing the largest explicit-tile effect.

By contrast, iTS-HVox showed a substantially stronger pseudo-voxel deletion effect. Deleting the highest-attention pseudo-voxels produced a much larger hotspot-embedding perturbation than random deletion at all tested deletion fractions. The top-minus-random hotspot-embedding difference increased from approximately 0.35 to 0.56 from 5% to 30% deletion, far exceeding the explicit-tile S1-S3 effects. Together, these findings show that B3iTS-HVox learns a selective and retrieval-relevant hotspot representation: high-attention pseudo-voxels are necessary for preserving the hotspot embedding and its nearest-neighbour retrieval structure, whereas random and low-attention pseudo-voxels have substantially weaker effects.

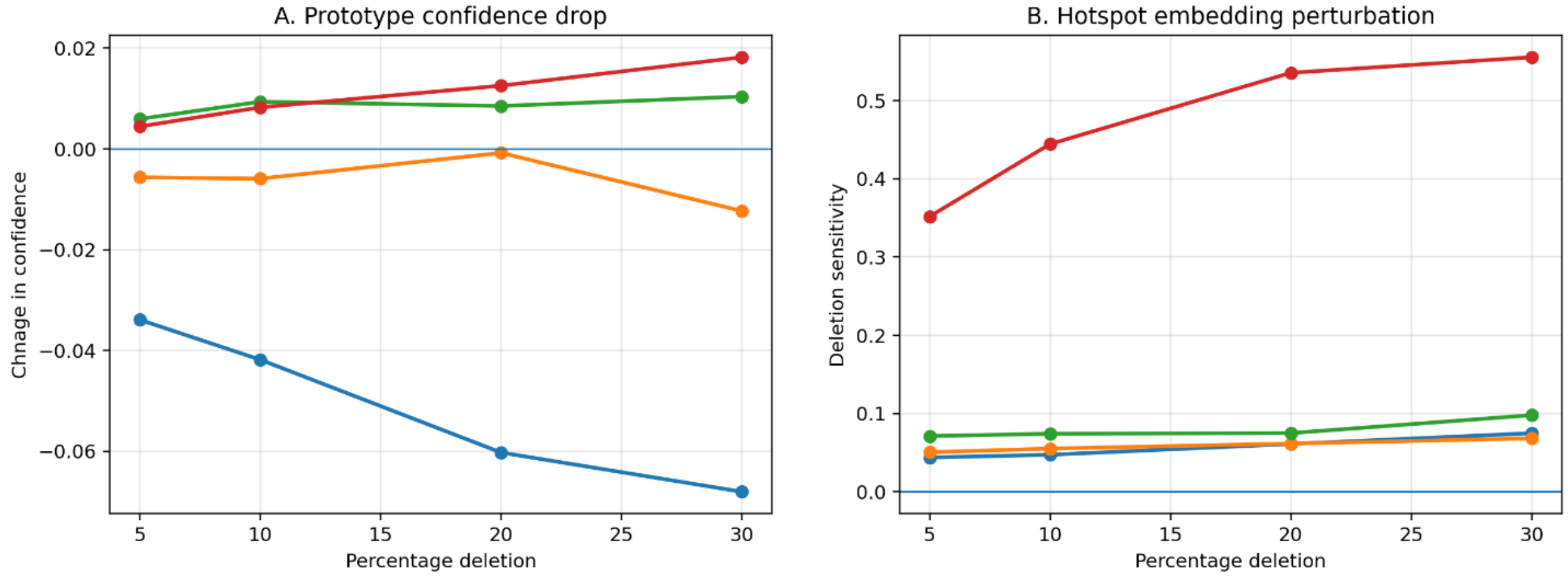

Figure 4. Top attention-deletion effect relative to random deletion across experiments. The y-axis shows the difference between high-attention deletion and random deletion, computed as Top − Random, across deletion fractions of 5%, 10%, 20% and 30%. Plot (A) shows the change in prototype confidence, and (B) shows hotspot-embedding perturbation. Blue, orange and green line represent ablations S1, S2, S3 and red represents iTS-HVox method.

### 3.3. Qualitative visualisation

The qualitative iTS-HVox visualisation in Figure 5 illustrates how the model maps PET-derived tumour signal into spatially organised pseudo-voxel phenotype prototypes. Images in the left column (A, D, G) show the PET-CT scan identifying the tumour location within the body in three example patients. The middle column (B, E, H) shows the original PET tumour uptake values, where higher values are concentrated along the central tumour segment, reflecting heterogeneous metabolic activity within the lesion. The right column (C, F, I) shows the same tumour volume labelled by the optimal-transport-informed pseudo-voxel prototype assignment, with most voxels assigned to the tumour-aggressiveness prototype, interspersed with smaller regions corresponding to tumour instability and tumour suppression. This spatially heterogeneous prototype distribution suggests that the model does not treat the lesion as a uniform PET volume but instead decomposes it into biologically interpretable regions.

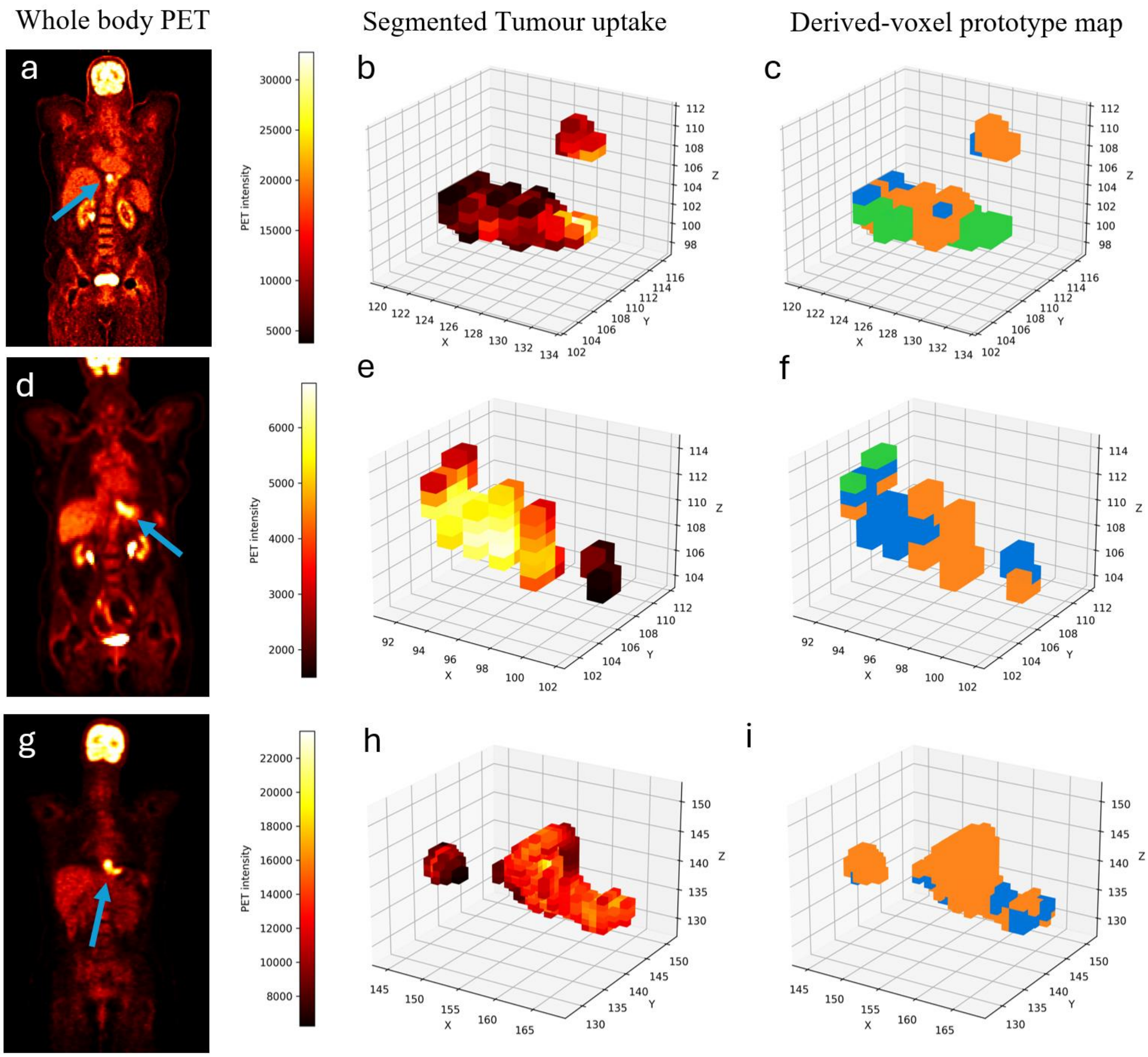

**Figure 5.** Qualitative visualisation of iTS-HVox PET derived-voxel habitats for three patients. Representative case showing the original whole body PET with tumor highlighted with arrow (left), segmented tumour uptake distribution (center) and the corresponding iTS-HVox optimal-transport-informed derived-voxel prototype map (right). PET intensity highlights heterogeneous metabolic uptake within the tumour, while the painted prototype map assigns derived-voxels to H&E-derived phenotype prototypes, including tumour aggressiveness (orange), instability (blue) and suppression (green).

## 3.4 Tumour volume-clinical stage concordance

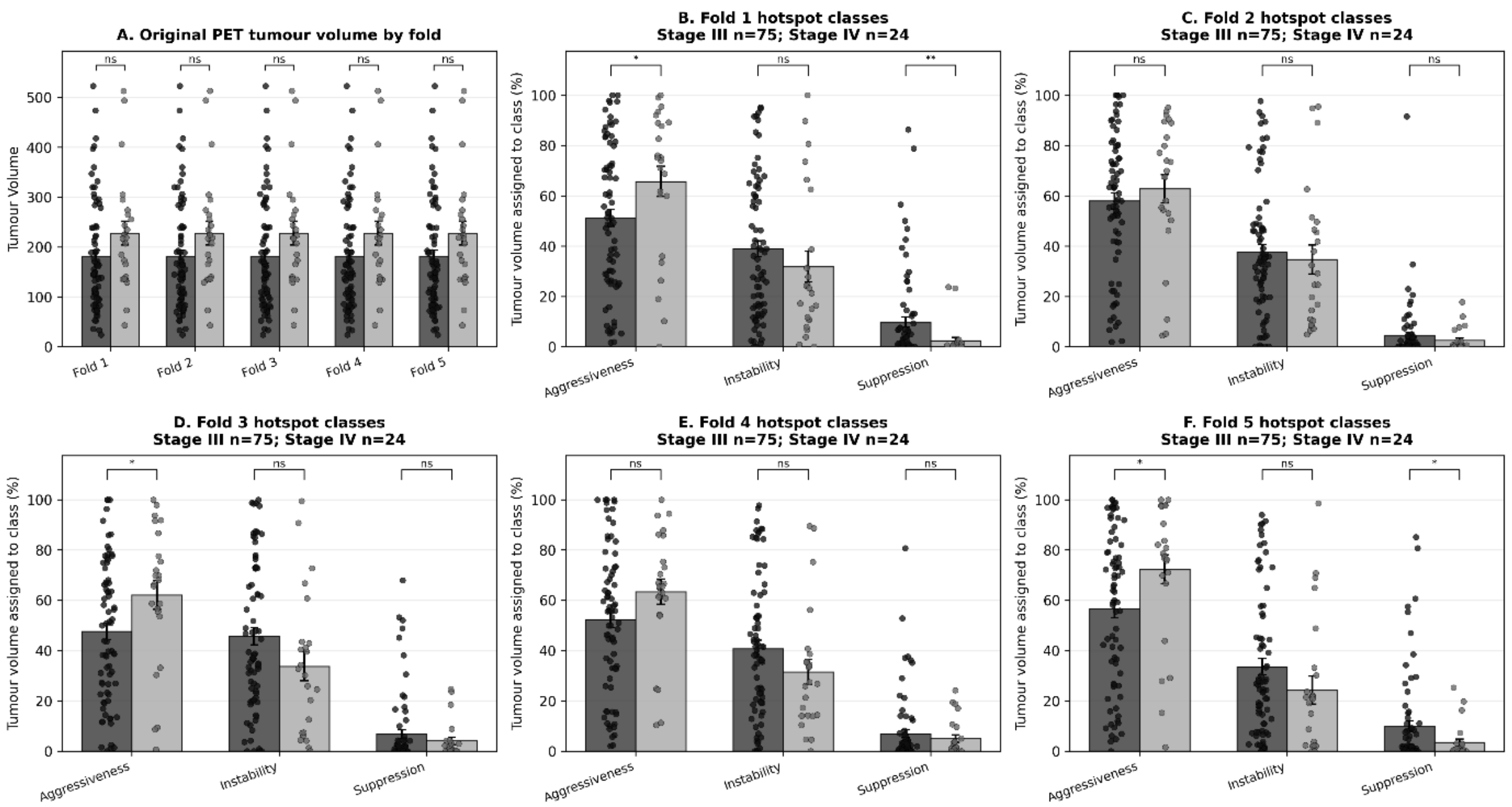

Figure 6 Relative distribution of PET-derived tumour habitats differs between clinical stages. A) across stage III and stage IV patients total tumour volume. B–F) compares H&E derived class phenotypes with respect to the total tumor volume in stage III and stage IV across folds. Significance labels were annotated as: $p<0.05$, *; $p<0.01$, **; $p<0.001$, ***; and non-significant comparisons were labelled ns.

Figure 6A showed that original PET tumour volume was broadly different between Stage III and Stage IV patients across all five folds. Figure 6 B-F shows the fold-wise hotspot analysis showed a recurring pattern in the distribution of iTS-HVox-derived tumour habitats. Across folds, the aggressiveness prototype occupied a larger proportion of tumour volume in Stage IV patients than in Stage III patients. This difference was statistical significance in Fold 1, Fold 3 and Fold 5. This suggests that the PET-derived aggressiveness habitat was more prominent in advanced-stage disease. In contrast, the instability prototype generally showed either non-significant differences or a tendency toward higher representation in Stage III patients. Similarly, the suppression prototype occupied a smaller fraction of tumour volume overall and showed fold-dependent behaviour. In some folds, suppression was more represented in Stage III patients, with statistically significant differences visible in Fold 1 and Fold 5, whereas other folds showed non-significant differences.

## 4. Discussion

This study investigated a framework for PET representation learning for content-based medical image retrieval under weak multimodal supervision. Rather than treating PET retrieval as a whole-lesion feature-matching problem, the proposed framework progressively introduced hotspot conditioning, PET–H&E alignment and implicit derived-voxel supervision to learn representations for intra-tumour heterogeneity. The central premise was that $^{18}$F-FDG uptake measured using PET contains clinically meaningful metabolic structure beyond global uptake summaries, but they are difficult to capture using patient-level labels alone. By incorporating H&E-derived weak supervision during training while preserving PET-only inference through student distillation, the framework aimed to bridge macroscopic functional imaging with microscopic tumour morphology without requiring spatial PET–histology registration.

The results can be interpreted as the learned hotspot representation was functionally faithful to a sparse subset of PET pseudo-voxel evidence, rather than being driven by diffuse lesion-wide information. Retrieval results in Figure 2 highlighted that hotspot-conditioned embeddings were not equivalent to global PET embeddings. Across the explicit-tile and implicit pseudo-voxel experiments, hotspot and global representations behaved differently under retrieval and perturbation analyses. The attention-deletion experiments provided strong evidence for this distinction. As could be deduced from Figures 3 and 4, deleting the highest-attention iTS-HVox pseudo-voxels caused a markedly larger perturbation of the hotspot embedding than deleting random or low-attention pseudo-voxels. The top-attention deletion effect was approximately 9–40 times larger than random deletion depending on the deletion fraction, while low-attention deletion produced minimal hotspot-embedding change.

The iTS-HVox results also clarified the role of implicit pseudo-voxel supervision. By processing the 3D PET tumour crop through a convolutional encoder, the model avoided dependence on manually defined explicit PET tiles and instead used downstream feature-grid locations as pseudo-voxels. This allowed optimal-transport supervision to act at a local feature-grid level while still exporting a PET-only retrieval embedding. The strong attention-deletion response of iTS-HVox indicated that this implicit representation could learn retrieval-relevant local evidence. However, the result should be interpreted carefully. The strongest validation was observed for hotspot embedding perturbation and retrieval stability, whereas global embedding perturbation behaved differently. In particular, low-attention deletion affected the global embedding more than the hotspot embedding at higher deletion fractions, which is expected because the global embedding aggregates broader lesion context. Thus, B3iTS-HVox improves the faithfulness and selectivity of the hotspot representation, not that all PET representations are equally hotspot-dependent.

The ablation ladder as summarised in Table 1 is important from a pattern-recognition perspective because it shows that the performance gain is not simply due to adding more supervision but depends on how weak multimodal information is structured and transferred.

Table 1. Ablation progression and rationale.

| Strategy | Architecture | H&E use during training | Rationale | Downstream analysis |
|---|---|---|---|---|
| S1 | Explicit tile teacher-student with PET/H&E token interface; no H&E-derived PET-tile pseudo-label loss. | H&E may inform the multimodal teacher, but PET-tile pseudo-label transfer is disabled. | Establishes whether PET hotspot distillation alone is useful. | Hotspot-conditioned PET retrieval versus global PET retrieval. |
| S2 | Same explicit tile architecture as S1. | Nearest H&E weak label is transferred to PET tiles using raw PET-H&E embedding similarity. | Tests the naive assumption that equal embedding dimension implies a valid shared space. | Benefit and failure mode of raw cross-modal pseudo-labelling. |

| S3 | Explicit tile architecture with fold-specific aligned PET-H&E latent space. | Pseudo-label transfer is performed after train-fold PET-H&E alignment. | Reduces raw-space mismatch before H&E-to-PET transfer. | Value of aligned cross-modal weak supervision. |
|---|---|---|---|---|
| iTS-HVox | Implicit 3D CNN PET encoder with phenotype prototype maps and optimal-transport pseudo-voxel supervision. | Labelled H&E patches are softly transported to PET feature-grid locations. | Avoids hard nearest-neighbour pseudo-labels and learns spatially organised PET prototypes. | Preferred downstream method for PET-only retrieval and hotspot interpretation. |

These findings highlight the need to evaluate learned attention beyond visual plausibility. The framework separates four questions that are often conflated in medical image representation learning: whether a representation improves retrieval, whether hotspot and global embeddings encode different information, whether high-attention regions are functionally necessary for the embedding, and whether the representation is concordant with independent biological phenotypes. The tumour-differentiation retrieval results are informative because the retrieval query at inference used only PET images, whereas the tumour-differentiation label was derived from H&E histopathology. Therefore, the observed retrieval performance suggests that PET embeddings retain a measurable, although indirect, signal related to histological differentiation. The strongest performance was generally observed for the B3its-HVox hotspot representation, particularly for early-rank metrics such as *mAP@*3 and *MRR@k*, indicating that the implicit pseudo-voxel hotspot embedding was better able to retrieve patients with similar H&E-defined tumour differentiation than global PET embeddings or generic PET-MIP deep features.

Recent radiomics and PET/CT AI studies continue to support the central premise that voxel-intensity heterogeneity in oncologic PET contains prognostic and biological information beyond simple tumour size or SUV summaries [25]. In oesophageal cancer, earlier H&E [26] and PET radiomics work showed that baseline intratumoural heterogeneity and temporal PET texture changes could predict pathological response and survival after chemoradiotherapy, and spatial–temporal FDG-PET features were also explored for modelling pathological response. For example, [27]. In another study and [28]. These studies align with our finding that PET contains clinically meaningful heterogeneity beyond global tumour burden. However, our results differ in how this heterogeneity is represented. Conventional radiomics summarises the tumour using predefined histogram, shape and texture descriptors, whereas iTS-HVox decomposes the PET tumour into learned model-derived habitats and quantifies the relative tumour volume assigned to aggressiveness, instability and suppression prototypes. In Figure 6 overall PET tumour volume was not consistently different between Stage III and Stage IV, but the aggressiveness prototype occupied a larger fraction of tumour volume in Stage IV across several folds [29]. This suggests that the learned PET habitat composition captures a stage-associated metabolic phenotype that is not simply explained by tumour size, extending prior radiomics findings from handcrafted global heterogeneity features toward H&E-informed, PET-derived phenotype distributions available at PET-only inference.

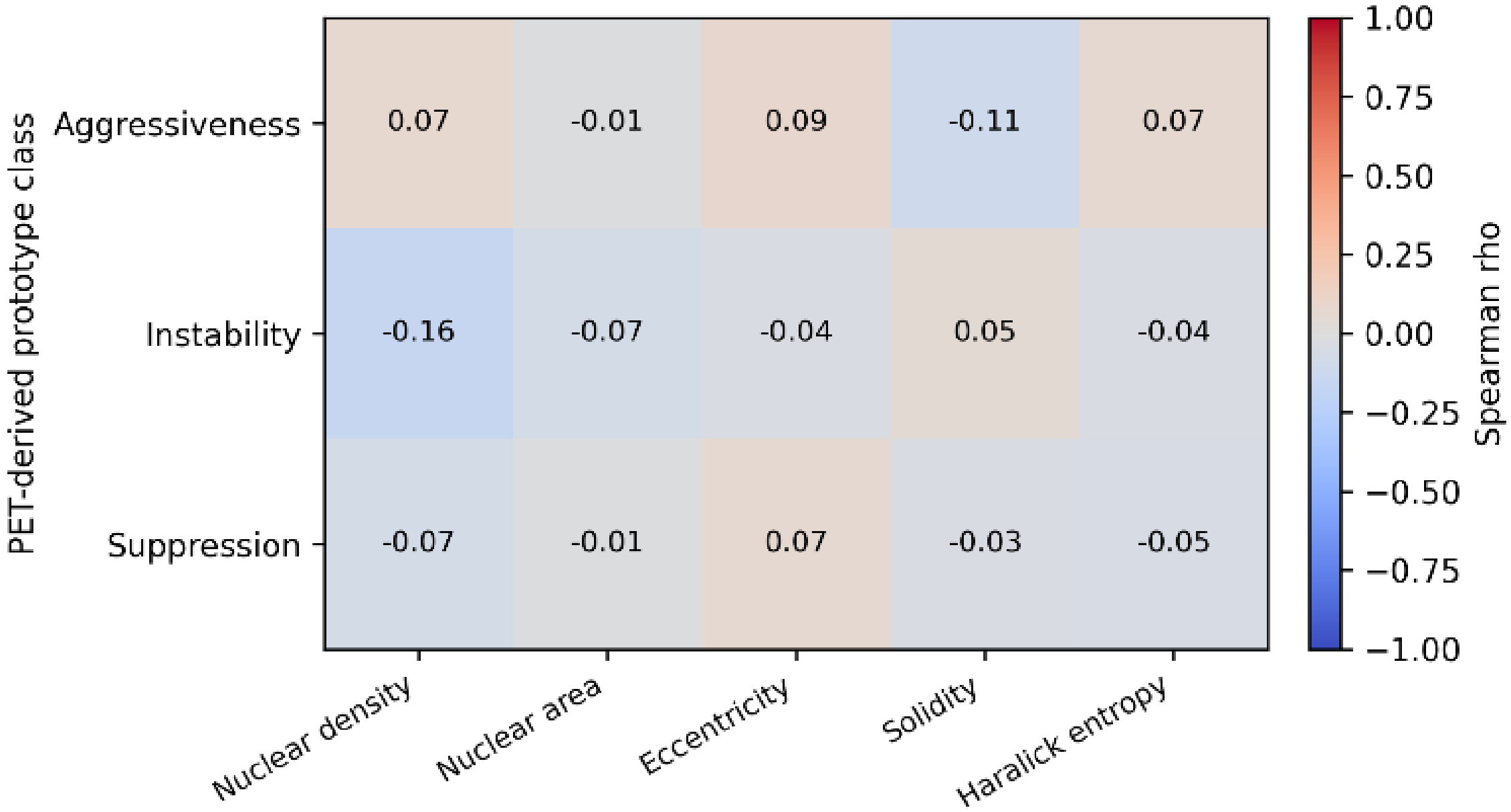

Figure 7: Concordance of iTS-HVox prototype classes using histopathology features and original PET uptake intensities.

Figure 7 shows across patients prototype-specific PET uptake enrichment showed weak associations with the selected histomic features. The aggressiveness prototype showed small positive associations with nuclear density, eccentricity and Haralick entropy, and a weak negative association with solidity, a pattern partly consistent with increased nuclear irregularity and texture heterogeneity. The histomics concordance analysis in Figure 7 showed biologically interpretation of the iTS-HVox prototype classes. Because PET and H&E were not spatially registered, these analyses should not be viewed as evidence that PET hotspots localise specific microscopic regions. Instead, they test whether voxel-derived PET prototype behaviour shows patient-level concordance with interpretable nuclear morphology. It is consistent with prior PET literature showing that nuclear size, shape and texture features capture malignant cytological atypia and can relate to tumour grade or prognosis [30].

## 5. Conclusion

We demonstrated that PET-only tumour retrieval can be improved by learning hotspot-conditioned representations that emphasise biologically informative regions rather than relying solely on whole-tumour/global image descriptors. These findings support the feasibility of PET-based virtual biopsy as a hypothesis-generating strategy for identifying tumour habitats, while also emphasising that clinical translation will require prospective validation against targeted biopsy, spatial pathology, molecular profiling and multi-centre imaging cohorts.

Hence, our framework provides a future direction for PET-based virtual biopsy as a non-invasive approach for identifying biologically informative tumour habitats in cancer care by verifying whether PET-derived hotspots correspond to clinically meaningful biological habitats, such as poorly differentiated tumour regions, hypoxic or metabolically aggressive sub volumes, immune-excluded microenvironments, or treatment-resistant clones.

Importantly, validation must move beyond retrospective retrieval performance to clinically actionable endpoints: improved biopsy yield, better risk stratification, prediction of treatment response, and reproducibility across scanners, centres and tumour types. If confirmed, this would support a clinically useful role for non-invasive PET diagnostics not as a replacement for tissue pathology, but as a decision-support tool to guide where tissue should be sampled, how tumour heterogeneity should be interpreted, and how imaging-derived biological habitats can be incorporated into precision oncology workflows.

## 5. Acknowledgement

The authors wish to thank the patients that have participated in our research project. We acknowledge the work and support of the Upper GI Unit at the Princess Alexandra Hospital, Brisbane, Australia. This research was partially carried out at the Translational Research Institute, Woolloongabba, QLD 4102, Australia. The Translational Research Institute is supported by a grant from the Australian Government. This study was funded by a large Cancer Council Queensland (CCQ) grant, aimed at Accelerating Collaborative Cancer Research (*ACCR-190*).